\documentclass[10pt,twocolumn,letterpaper]{article}

\usepackage{cvpr}              % To produce the CAMERA-READY version
\usepackage{float}
\usepackage[linesnumbered,ruled,vlined]{algorithm2e}
\usepackage{algorithmic}
\usepackage{amsmath}
\usepackage{amssymb}
\usepackage{fontawesome5}
\usepackage{graphicx}
\usepackage{float}
\usepackage[table]{xcolor}
\usepackage{xcolor}
\usepackage{multirow}

\definecolor{cvprblue}{rgb}{0.21,0.49,0.74}
\usepackage[pagebackref,breaklinks,colorlinks,allcolors=cvprblue]{hyperref}

\def\paperID{} % *** Enter the Paper ID here
\def\confName{CVPR}
\def\confYear{2026}
\title{\textcolor{blue!50}{\faLightbulb}\ CUE: Concept-Aware Multi-Label Expansion to Mitigate Concept Confusion in Long-Tailed Learning}
\author{Ruichi Zhang$^{1}$\footnotemark[1]\hspace{20pt}Chikai Shang$^{1}$\footnotemark[1]\hspace{20pt}Jiacheng Yang$^{1}$\hspace{20pt}Mengke Li$^2$\\
Yang Zhou$^3$\hspace{20pt}Junlong Gao$^{1}$\hspace{20pt}Yang Lu$^{1}$\footnotemark[2]\\
{\small $^1$Key Laboratory of Multimedia Trusted Perception and Efficient Computing, Ministry of Education of China, Xiamen University, China}\\
{\small $^2$College of Computer Science and Software Engineering, Shenzhen University, China}\\
{\small $^3$Institute of High Performance Computing, A*STAR, Singapore}\\
{\tt\small zhangruichi@stu.xmu.edu.cn\hspace{20pt}ckshang12@gmail.com\hspace{20pt}23020250157840@stu.xmu.edu.cn}\\
{\tt\small mengkeli@szu.edu.cn\hspace{20pt}zhou\_yang@a-star.edu.sg\hspace{20pt}jlgao@xmu.edu.cn\hspace{20pt}luyang@xmu.edu.cn}
}

\begin{document}
\twocolumn[{%
\maketitle

\vspace{-2em}
\begin{center}
    \centering
    \begin{minipage}[b]{1.0\textwidth}
        \centering
        \includegraphics[width=0.95\linewidth]{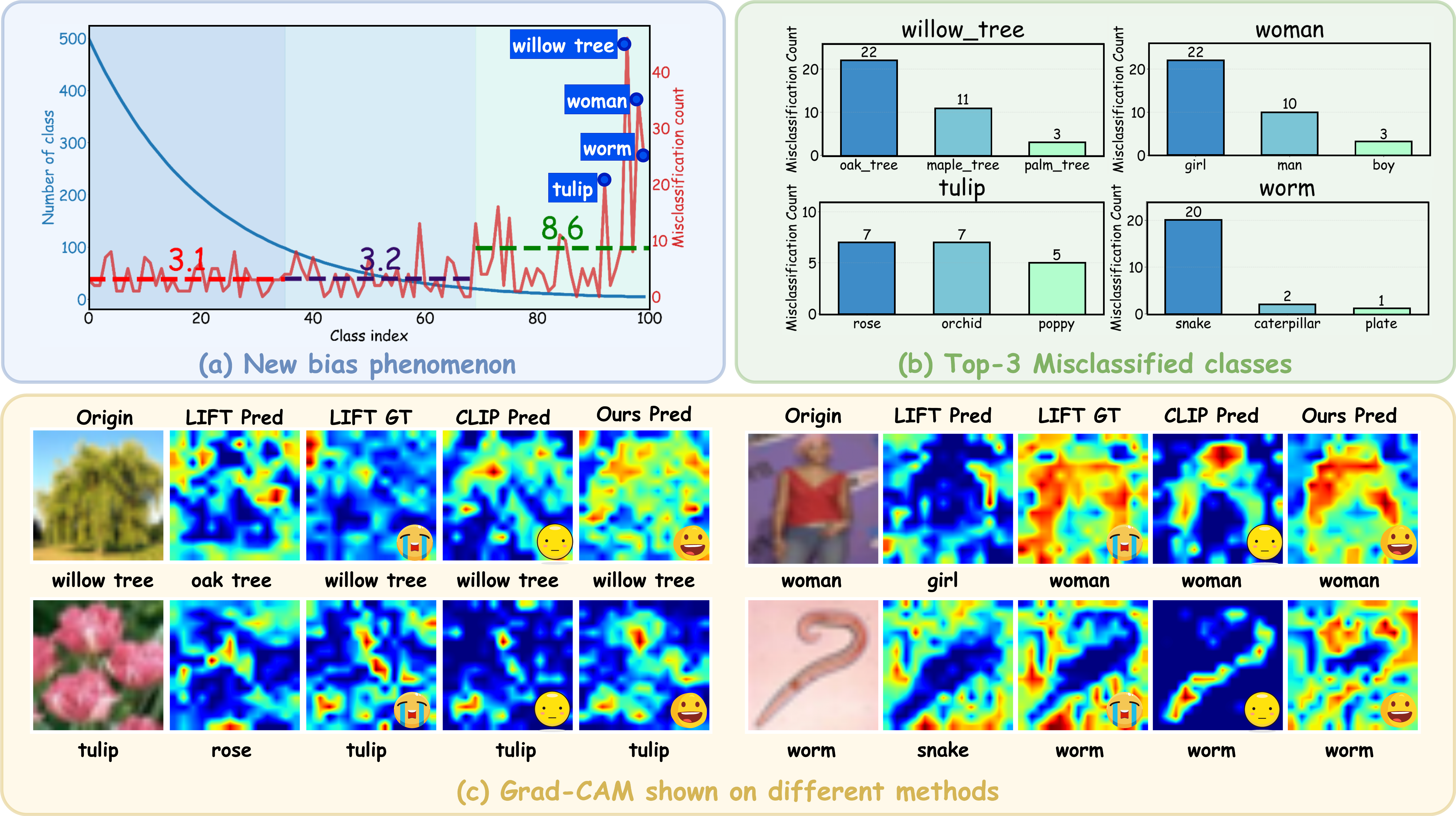}
        \vspace{-0.5em}
      \captionof{figure}{\textbf{Motivation: Handling Concept Confusion, Not Just Class Imbalance.} (a) On CIFAR100-LT (IR=100), fine-tuning CLIP with LIFT~\cite{shi2024long} turns many zero-shot correct predictions into errors, especially on tail classes. (b) Most of these errors are misclassified into semantically related categories, a phenomenon we term as \textit{concept confusion}. This confusion mainly arises from the fine-tuning process, which disrupts inter-class relationships among semantically similar concepts, as evidenced by attention shifts revealed in Grad-CAM visualizations (c). These observations motivate exploring strategies to preserve inter-class relationships during fine-tuning.
      }
      % \captionof{figure}{}
        \label{fig:motivation2}
    \end{minipage}
\end{center}
}]

% \maketitle

\footnotetext[1]{Equal contribution.}
\footnotetext[2]{Corresponding author: Yang Lu (luyang@xmu.edu.cn).}
\footnotetext[3]{Accepted by CVPR 2026.}

\begin{abstract}

\vspace{-1em}
Long-tailed distributions are common in real-world recognition tasks, where a few head classes have many samples while most tail classes have very few. Recently, fine-tuning foundation models for long-tailed learning has gained attention due to their excellent performance. However, most existing methods focus solely on mitigating long-tailed distribution bias while overlooking concept confusion caused by the long-tailed distribution. In this paper, we study this problem and attribute it to the mutual exclusivity of single-label supervision under long-tailed distributions, which suppresses feature sharing among related classes and amplifies the dominance of head classes, leading to disrupted inter-class discriminability. To address this, we propose \textbf{CUE},  \underline{C}oncept-aware m\underline{U}lti-label \underline{E}xpansion, which introduces multi-label concept signals to preserve disrupted inter-class relationships. Specifically, CUE constructs concept sets by \textbf{(i)} extracting instance-level visual cues from zero-shot CLIP and \textbf{(ii)} generating class-level semantic cues with LLM; the two cues are incorporated via separately weighted Binary Logit-Adjustment (BLA) auxiliary losses and jointly optimized with the baseline Logit-Adjustment (LA) loss. Experiments on several long-tailed benchmarks, CUE achieves balanced and strong performance, surpassing recent state-of-the-art methods. 
% The code is available in the supplementary materials.
\textit{Code is available at:} \url{https://github.com/zhangruichi/CUE}.

\end{abstract}    
\vspace{-5mm}
\section{Introduction}
\label{sec:intro}

% 原版
% In real-world classification problems, data often exhibit a long-tailed label distribution, where a few ``head classes'' have abundant samples, while a large number of ``tail classes'' contain only a few training samples. This problem, known as long-tailed learning, aims to overcome this imbalance by improving recognition performance on tail categories without compromising accuracy on head classes. It has been extensively studied through data-level, representation-level, and model-level strategies, including re-sampling, re-weighting, decoupled training, and contrastive representation learning. These approaches have laid the foundation of long-tailed learning and achieved remarkable progress in mitigating long-tailed bias in classification.

% v0
% In real-world classification tasks, data naturally exhibits a long-tailed distribution over classes, where a few head classes dominate with abundant samples while most tail classes contain only a few samples~\cite{van2018inaturalist, zhang2023deep, zhang2025systematic}. This severe imbalance poses a fundamental challenge for standard learning approaches, which often struggle to generalize to tail classes and tend to be heavily biased toward head classes~\cite{menon2021longtail, li2022long}. To addre ss this, the field of long-tailed learning (LTL) has been extensively studied to mitigate this challenge and achieve more balanced performance~\cite{cao2019learning, kang2020decoupling, zhang2021distribution, hou2025learning}.

% v1
In real-world classification tasks, data often follow a long-tailed distribution, where a few head classes have abundant samples (e.g., frequent animals in image datasets or common diseases in medical datasets), while numerous tail classes have only limited examples (e.g., rare animal species or uncommon diseases)~\cite{zhang2023deep, zhang2025systematic, johnson2019mimic}. This imbalance is prevalent in applications including image recognition, medical diagnosis, and autonomous driving~\cite{van2018inaturalist, yasunobu2003auto, ao2020application}, often resulting in models that perform well in head classes but poorly in tail ones~\cite{menon2021longtail, li2022long}. To address this issue, the field of long-tailed learning (LTL) has emerged, aiming to develop methods that achieve balanced and promising performance across all classes~\cite{cao2019learning, kang2020decoupling, hou2025learning}.

Recently, fine-tuning foundation models for long-tailed learning has gained considerable attention due to their strong generalization and excellent performance. Representative methods~\cite{ shi2024long, tian2022vl, long2022retrieval,li2024improving,xia2024lmpt} adapt foundation models through parameter-efficient fine-tuning such as prompt tuning~\cite{jia2022visual, dong2023lpt}, adapter tuning~\cite{jia2021scaling, steitz2024adapters}, or LoRA~\cite{hu2022lora}, and even by leveraging additional data to enhance tail performance. By transferring rich pre-trained knowledge, these models effectively enhance representation quality and improve recognition for tail classes, making them a promising paradigm for addressing long-tailed challenges.

However, most existing methods focus solely on mitigating the bias from well-studied class imbalance while overlooking that from \textit{concept confusion} that arises during fine-tuning. As shown in Fig.~\ref{fig:motivation2}(a), many samples correctly classified by zero-shot CLIP become misclassified after fine-tuning, particularly for tail classes. These errors are frequently assigned to semantically related categories, as illustrated in Fig.~\ref{fig:motivation2}(b). Grad-CAM visualizations in Fig.~\ref{fig:motivation2}(c) further show that fine-tuning shifts the model’s attention from object-centric regions to incorrect areas, resulting in semantically inconsistent predictions for tail classes. Although the zero-shot CLIP is less adapted to the target domain, it still maintains relatively coherent attention, indicating that this confusion primarily emerges during fine-tuning. These observations indicate that in addition to class imbalance, \textit{concept confusion} has become a key challenge in adapting foundation models to long-tailed distributions.

In this paper, we attribute this concept confusion to the mutual exclusivity inherent in single-label supervision, which enforces each instance to belong to only one class even when categories are semantically or visually related. Under long-tailed distributions, this exclusivity encourages the model to favor classes with more abundant and representative samples in pursuit of better training objectives, causing these head classes to dominate the learning process and disrupt the correlation among related categories. Consequently, inter-class relationships are disrupted, which further amplifies concept confusion.

To alleviate this issue, we propose CUE (\underline{C}oncept-aware m\underline{U}lti-label \underline{E}xpansion), which encourages the model to actively seek informative cues.  CUE introduces the strengths of vision–language models (VLMs) and large language models (LLMs) as guidance to generate multiple semantically related class labels, enabling the model to explicitly learn inter-class relationships rather than collapsing them into single-label decisions. Specifically, VLMs provide instance-level cues, capturing fine-grained visual and contextual relationships through large-scale image–text pre-training~\cite{radford2021learning,jia2021scaling}, while LLMs offer class-level cues, modeling higher-level semantic and conceptual associations among categories~\cite{brown2020language,bommasani2021opportunities}. By integrating these cues, CUE effectively preserves the inter-class structure learned during pre-training and guides the model to make more balanced predictions across all classes, while also producing more accurate and object-centric attention, as illustrated in Fig.~\ref{fig:motivation2}(c). Our contributions are threefold:
\vspace{0.5em}
\begin{itemize}
    \item We identify and analyze \textit{concept confusion}, that arises during fine-tuning of foundation models on long-tailed data, and reveal its underlying cause rooted in the mutual exclusivity of single-label supervision.
    \item We propose \textbf{CUE}, which integrates VLMs and LLMs to generate semantically related labels and explicitly model inter-class relationships, with a plug-and-play design that flexibly supports both fine-tuning and from-scratch.
    \item We demonstrate that \textbf{CUE} effectively mitigates concept confusion and achieves consistent gains on multiple long-tailed benchmarks.  
\end{itemize}

% \item 
% We develop a scalable confusable-class generation pipeline that constructs high-quality semantic similarity sets through batched generation and filtering.
% We develop a confusable-class generation pipeline that constructs high-quality semantic similarity sets through batched generation and filtering.
% v2 

\section{Related Work}
\label{sec:related}

\noindent \textbf{Long-Tailed Learning with Foundation Models.}
Foundation models have recently demonstrated remarkable effectiveness in LTL by leveraging their rich pre-trained knowledge to enhance long-tailed performance~\cite{tian2022vl, dong2023lpt, wang2024exploring, shi2025lift+,zhang2024long,luo2025long,lt2026hong}. A common approach is to combine various LTL strategies~\cite{cui2019class,zhang2021distribution, menon2021longtail,hong2021disentangling,shi2023re} with fine-tuning. For instance, LPT~\cite{dong2023lpt} integrates multiple long-tailed tricks (e.g., re-weighting, decoupling, and cosine classifier) with prompt tuning~\cite{jia2022visual}. Similarly, LIFT~\cite{shi2024long} adopts the balanced LA loss~\cite{menon2021longtail} under the lightweight fine-tuning paradigm. The central goal of these methods is to adapt upstream knowledge to downstream tasks while mitigating the bias introduced by class imbalance. However, beyond data-introduced bias, we observe that fine-tuning itself introduces \textit{concept confusion}, i.e., tail-class samples are easily confused with semantically related categories after fine-tuning, which has been largely overlooked. This paper, in contrast, aims to leverage the strong priors within foundation models to alleviate such bias.
% However, distinct from bias incurred by data, this paper reveals that fine-tuning introduces a new form of bias (even with the advanced LA loss): many samples that were correctly classified by zero-shot foundation models become misclassified after fine-tuning, particularly for tail classes.
% It can be specifically divided into two categories: (1) Directly fine-tuning on foundation models. (2) Introducing additional priors from foundation models.
% \vspace{0.5em}

\noindent \textbf{Pseudo-Labeling.}
% Pseudo-labeling mainly aims to generate artificial labels for unlabeled data and train the models in a self-training manner. For instance, FixMatch employs a fixed confidence threshold to select high-quality pseudo-labels. Similarly, SoftMatch adopts a soft threshold to exploit both high-quality and high-quantity pseudo-labels. These approaches, specifically, aim to generate a single pseudo-label for each unlabeled data to harness the unlabeled samples. Distinct from these approaches, this paper aims to leverage the strong priors of foundation models to generate multiple pseudo-labels, aiming to fully exploit the labeled samples to better learn inter-class correlation.
Pseudo-labeling primarily aims to generate artificial labels for unlabeled samples to better exploit unlabeled data~\cite{hu2021simple, wang2022debiased, abdelfattah2023cdul}. For instance, FixMatch~\cite{sohn2020fixmatch} uses a fixed confidence threshold to select high-quality pseudo-labels for self-training. SoftMatch~\cite{chen2023softmatch}, on the other hand, employs a soft threshold to balance quality and quantity of pseudo-labels. OTAMatch~\cite{zhang2024otamatch} further advances this paradigm by formulating pseudo-labeling as an optimal transport assignment process. The central goal of these approaches is to generate one pseudo-label for each unlabeled sample for supervised learning. In contrast, existing pseudo-labeling methods do not focus on long-tail learning, and they aim to infer a single ground-truth label, whereas we instead seek multiple labels to identify semantically related categories.
% \vspace{0.5em}

\noindent \textbf{Parameter-Efficient Fine-Tuning.}
This paper mainly focuses on parameter-efficient fine-tuning (PEFT)~\cite{xin2024parameter, mai2025lessons, fu2025dtl}, as Shi \textit{et al.}~\cite{shi2024long} demonstrate that full fine-tuning can significantly degrade performance on tail classes, whereas lightweight fine-tuning better preserves upstream knowledge and benefits them. Specifically, PEFT approaches can be grouped into three categories: (i) reparameter tuning (e.g., LoRA~\cite{hu2022lora}), which integrates newly trainable parameters into the pre-trained weights; (ii) prompt tuning (e.g., VPT~\cite{jia2022visual}), which injects learnable prompt tokens into the input space of Transformer blocks; and (iii) adapter tuning (e.g., Adapter~\cite{houlsby2019parameter} and Adaptformer~\cite{chen2022adaptformer}), which inserts extra learnable MLP modules within each Transformer block.

\section{Proposed Method: CUE}
\label{sec:method}
\begin{figure*}[t]
  \centering
  % \vspace{-1em}
  \includegraphics[width=\textwidth]{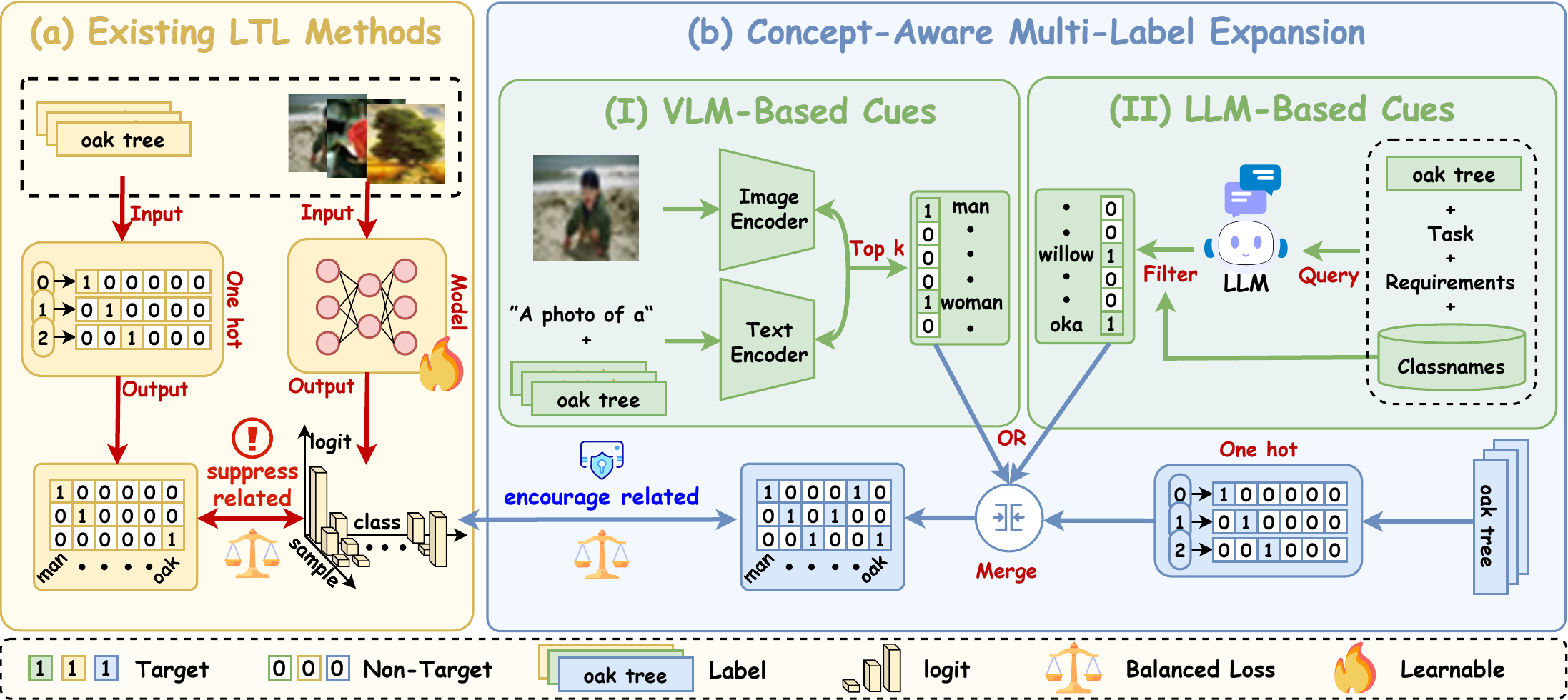}
  \vspace{-1.6em}
    \caption{\textbf{Overall framework.}
    This figure illustrates the overall structure of our approach compared with the existing long-tailed learning methods.
    (a) \textit{Existing LTL methods} rely on a single-label balanced loss, which tends to suppress semantically related categories and bias learning toward head classes, often leading to concept confusion.
    (b) \textit{CUE} augments the main balanced loss with two concept-aware cues: (I) VLM-based instance-level cues and (II) LLM-based class-level cues. By jointly optimizing these cues with the BLA objective, CUE encourages learning from semantically related categories rather than suppressing them, effectively alleviating concept confusion.}
    \label{fig:overview}
    \vspace{-1.3em}
\end{figure*}

As discussed in Sec.~\ref{sec:intro}, fine-tuning foundation models on long-tailed data often amplifies concept confusion. We attribute this phenomenon to the exclusivity of single-label supervision, which suppresses the shared representations among semantically related categories and distorts the inter-class relationships established during pre-training. To address this issue, our framework, shown as Fig.~\ref{fig:overview}, extends the typical long-tailed learning paradigm by adding a plug-and-play module, CUE, which explicitly mitigates such confusion through joint supervision from VLM-based and LLM-based cues. These cues provide complementary visual and semantic guidance, enabling the model to preserve the pre-trained knowledge while adapting to the new  domain during fine-tuning. 
% The detailed method process is summarized in Algorithm~\textcolor[HTML]{367DBD}{1} in the Appendix~\textcolor[HTML]{367DBD}{A}.

\subsection{Preliminaries}
\label{sec:prelim}
\noindent \textbf{Notation.} We denote the training set with $N$ samples as $\mathcal{D}=\{(\mathbf{x}_i, y_i)\}_{i=1}^N$, where $\mathbf{x}_i$ indicates the $i$-th image sample and $y_i$ denotes its corresponding label. 
Let $n_c$ represent the number of samples in class $c$, and $C$ be the total number of classes. Given an input image $\mathbf{x}$, the model outputs a logit vector $\boldsymbol{\theta}(\mathbf{x}) \in \mathbb{R}^C$ corresponding to $C$ categories, where $\theta_y(\mathbf{x})$ denotes the logit corresponding to the label $y$.
The empirical class prior is given by $\pi_c = n_c / \sum_j n_j$.  
% \vspace{0.5em}

\noindent \textbf{Logit Adjustment.}
Logit Adjustment~\cite{menon2021longtail} rebalances long-tailed data by correcting the class posterior through Bayes’ rule. 
Since $\mathbb{P}(y|\mathbf{x})\!\propto\!\mathbb{P}(\mathbf{x}|y)\mathbb{P}(y)$, standard training on imbalanced data implicitly optimizes $\mathbb{P}_{\text{train}}(y|\mathbf{x})\!\propto\!\mathbb{P}(\mathbf{x}|y)\mathbb{P}_{\text{train}}(y)$, 
which biases predictions toward frequent classes. 
Assuming $\mathbb{P}(\mathbf{x}|y)$ remains unchanged between training and test distributions, 
a balanced posterior can be approximated as 
$\mathbb{P}_{\text{bal}}(y|\mathbf{x})\!\propto\!\mathbb{P}_{\text{train}}(y|\mathbf{x})/\mathbb{P}_{\text{train}}(y)
\!\propto\!\mathrm{softmax}(\theta_y(\mathbf{x})-\log\mathbb{P}_{\text{train}}(y))$.
Introducing a temperature $\tau$ yields the general form 
$\theta'_y(\mathbf{x})=\theta_y(\mathbf{x})-\tau\log\mathbb{P}_{\text{train}}(y)$, 
which can be applied during training to mitigate class imbalance. 
Since our framework adopts multi-label expansion, fair calibration between positive and negative samples is essential. 
To address this, we further extend the standard binary cross-entropy loss with same prior logit shift, 
resulting in a \emph{Binary Logit Adjustment (BLA)} formulation that aligns with our VLM- and LLM-based supervision. 
% A detailed derivation is provided in Appendix~\textcolor[HTML]{367DBD}{B}.

% In our framework, we start from a balanced single-label baseline (e.g., LA~\cite{menon2021longtail}) and extend it with \textbf{CUE}, which introduces multi-label supervision to capture visually and semantically confusable classes around the ground truth. 
% For fine-tuning scenarios, we employ a pre-trained CLIP backbone and freeze its parameters during training.
% ToDo: 这里根据情况 添加from scretch的内容
% ToDo: 是不是要加微调 以及 Logit adjustment 的相关Preliminaries

\subsection{VLM-Based Instance-Level Cues}
\label{sec:vlm_prior}
Considering that visual language model (VLM) naturally provides instance-level cues, we use their zero-shot predictions to capture potential cross-class confusion, thereby mitigating instance-level concept confusion. Specifically, for each training image $\mathbf{x}_i$, we compute its zero-shot prediction scores ${\theta}^{\text{zs}}(\mathbf{x}_i)$ by measuring the cosine similarities between the CLIP image embedding of $\mathbf{x}_i$ and the text prototypes generated from the pre-trained CLIP model using the standard prompt ``a photo of a [CLASS].'', and identify the Top-$k$ non-ground-truth categories that the model considers most similar:
\begin{equation}
\mathcal{T}^{\text{zs}}(x_i) = \text{Top-}k\Big(\operatorname*{argsort}_{y \neq y_i}\, {\theta}^{\text{zs}}_y(\mathbf{x}_i)\Big),
\end{equation}
where ${\theta}^{\text{zs}}_y(\mathbf{x}_i)$ denotes the zero-shot prediction logit of label $y$ obtained from the frozen CLIP model, and $\mathcal{T}^{\text{zs}}(\mathbf{x}_i)$ represents the Top-$k$ semantically similar non-ground-truth classes according to pre-trained embedding space of CLIP, which serve as additional visual and conceptual cues of $y_i$. 
To incorporate this relational cue into training, we augment the supervision signal by converting them into a binary multi-label target:
\begin{equation}
\tilde{t}^{\text{zs}}_{i} =
\begin{cases}
1, & c \in \{y_i\} \cup \mathcal{T}^{\text{zs}}(\mathbf{x}_i), \\
0, & \text{otherwise,}
\end{cases}
\end{equation}
where $\tilde{t}^{\text{zs}}_{i}$ denotes the enhanced multi-label supervision for sample $\mathbf{x}_i$.
These instance-level cues effectively preserve the local class neighborhood discovered during pre-training, maintaining the intrinsic visual structure learned by CLIP. It encourages the model to recognize visually correlated categories and maintain coherent representations across related classes, thereby mitigating concept confusion.

\subsection{LLM-Based Class-Level Cues}
\label{sec:llm_prior}

Similar to the instance-level cues derived from VLM, we further incorporate class-level semantic cues by leveraging large language model (LLM) to uncover relationships among original categories.
To mitigate concept confusion at the class level, we build for each class $c$ an offline neighbor set $\mathcal{N}^{\text{llm}}(c)$ using LLM. The LLM is prompted with the complete label vocabulary as the only candidate set and instructed to select semantic neighbors strictly from it. In practice, the LLM may still produce occasional out-of-vocabulary names, so we simply filter these out during post-processing to ensure all neighbors remain within the original label space.

\begin{figure}[h]
\centering
\includegraphics[width=\linewidth]{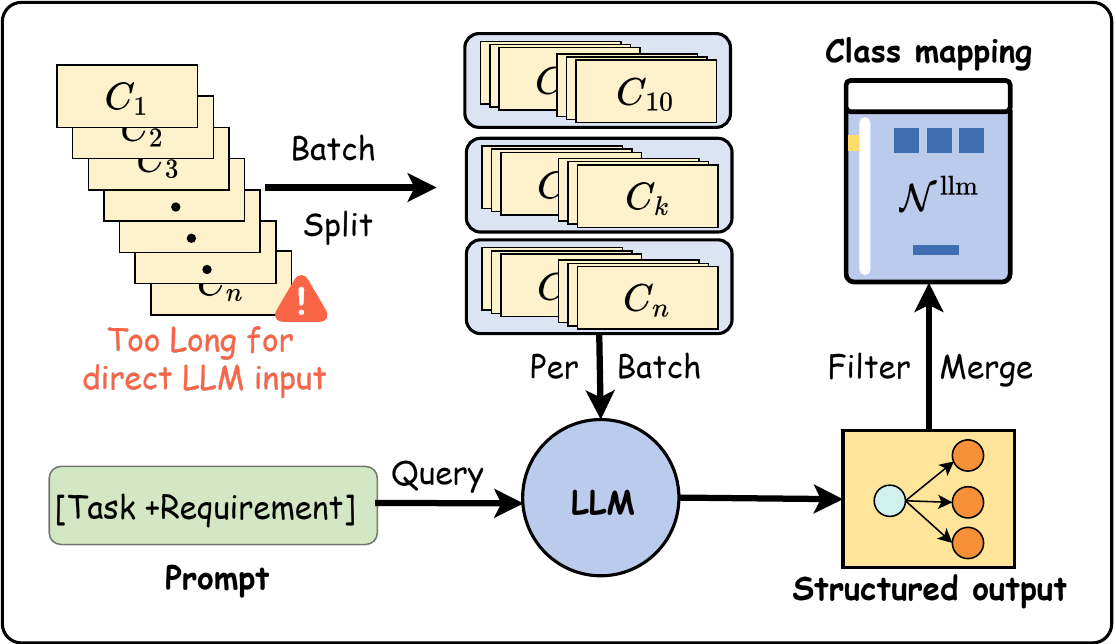}
\vspace{-1.0em}
\caption{\textbf{Construction pipeline of the LLM-based class-level cues}. This figure shows the process used to obtain the LLM-based semantic neighbor graph $\mathcal{N}^{\text{llm}}$.
}
\label{fig:llm_pipeline}
\vspace{-1.5em}
\end{figure}

However, directly prompting the entire label list tends to exceed the effective working range of the LLM, causing it to skip or collapse intermediate classes. As a result, the model exhibits degraded semantic quality, producing coarse or noisy relations instead of fine-grained confusion patterns. To ensure stable and comprehensive neighbor discovery across the full vocabulary, we design a \emph{batched prompting and filtering pipeline}: (i) the full label list is divided into batches, and each query is \emph{constrained} to return JSON mappings from class names to their nearest semantic neighbors strictly within the given subset; (ii) the batched outputs are merged and post-filtered to remove out-of-list, ambiguous, and duplicate entries, before aligning the class names to label indices (as shown in Fig.~\ref{fig:llm_pipeline}). Each batch follows a standardized instruction template to ensure consistent and structured LLM responses.

During training, each sample $(\mathbf{x}_i, y_i)$ receives additional class-level supervision constructed from its semantic neighbors:
\begin{equation}
\tilde{t}^{\text{llm}}_{i} =
\begin{cases}
1, & c \in \{y_i\} \cup \mathcal{N}^{\text{llm}}(y_i), \\
0, & \text{otherwise.}
\end{cases}
\end{equation}
where $\tilde{t}^{\text{llm}}_{i}$ denotes the augmented class-level multi-label target for sample $\mathbf{x}_i$, 
and $\mathcal{N}^{\text{llm}}(y_i)$ represents the set of neighbor classes of $y_i$ 
obtained from the LLM-generated class-level cues. These class-level cues capture broader semantic associations and complement the instance-level visual cues by introducing higher-level contextual structure.
They encourage the model to learn coherent representations across semantically related categories, thereby reducing concept confusion under long-tailed distributions.

% \begin{equation}
% \begin{split}
% \mathcal{L}
% = &~ \underbrace{\mathcal{L}^{\text{LA}}(x_i, y_i)}_{\text{baseline}} \\
% & + \lambda_{\text{zs}} 
%   \underbrace{\mathcal{L}^{\text{BLA}}(x_i, \tilde{\mathbf{t}}^{\text{zs}}_i)}_{\text{VLM prior}}
%   + \lambda_{\text{llm}} 
%   \underbrace{\mathcal{L}^{\text{BLA}}(x_i, \tilde{\mathbf{t}}^{\text{llm}}_i)}_{\text{LLM prior}},
% \end{split}
% \label{eq:overall_loss}
% \end{equation}

\subsection{Training Objectives}
\label{sec:objective}
Our training objective integrates the instance-level and class-level cues into a unified optimization framework that balances data-driven and cue-guided supervision. 
Specifically, we employ a single-label logit-adjusted (LA) loss and two multi-label binary logit-adjusted (BLA) terms corresponding to the VLM-based and LLM-based cues:
\begin{equation}
\begin{aligned}
\mathcal{L}
= &
~\underbrace{
  \colorbox{gray!15}{$\mathcal{L}^{\text{LA}}(\mathbf{x}_i, y_i)$}
}_{\text{baseline}}
+ \lambda_{\text{zs}}
  \underbrace{
  \colorbox{blue!10}{$\mathcal{L}^{\text{BLA}}(\mathbf{x}_i, \tilde{\mathbf{t}}^{\text{zs}}_i)$}
}_{\text{VLM cue}}
\\[2pt]
&+ \lambda_{\text{llm}}
  \underbrace{
  \colorbox{green!10}{$\mathcal{L}^{\text{BLA}}(\mathbf{x}_i, \tilde{\mathbf{t}}^{\text{llm}}_i)$}
}_{\text{LLM cue}},
\end{aligned}
\label{eq:overall_loss}
\end{equation}
where $\lambda_{\text{zs}}$ and $\lambda_{\text{llm}}$ are weighting coefficients for the two prior-guided terms, and 
$\tilde{\mathbf{t}}^{\text{zs}}_i$ and $\tilde{\mathbf{t}}^{\text{llm}}_i$ denote the multi-label supervision targets derived from the VLM-based instance-level and LLM-based class-level cues.

% \vspace{0.5em}

\noindent \textbf{Logit-Adjustment (LA).} The baseline loss adopts class-balanced logit adjustment, which introduces a log-prior shift with temperature~$\tau$:
\begin{equation}
\begin{split}
\theta'_c(\mathbf{x}) &= \theta_c(\mathbf{x}) + \tau \log \pi_c, \\
\mathcal{L}^{\text{LA}} 
&= -\log
\frac{\exp\!\big(\theta'_{y_i}(\mathbf{x}_i)\big)}
{\sum_{c}\exp\!\big(\theta'_{c}(\mathbf{x}_i)\big)},
\end{split}
\end{equation}
where $\pi_c$ denotes the empirical class prior defined in Sec.~\ref{sec:prelim}.
This formulation mitigates class imbalance by reducing the dominance of frequent classes through prior-dependent re-weighting.

% \vspace{0.5em}

\noindent \textbf{Binary Logit-Adjustment (BLA).}
For multi-label supervision, we apply a similar prior logit adjustment before the sigmoid activation:
\begin{equation}
\begin{split}
\tilde\theta_c(\mathbf{x}) &= \theta_c(\mathbf{x}) + \tau_{\text{b}}\log \pi_c, \\
\mathcal{L}^{\text{BLA}}
&= -\frac{1}{C}\sum_{c=1}^{C}\Big[
\tilde{t}_{i,c}\log\sigma\!\big(\tilde\theta_c(\mathbf{x}_i)\big) \\
&\qquad
+~(1-\tilde{t}_{i,c})\log\!\big(1-\sigma\!\big(\tilde\theta_c(\mathbf{x}_i)\big)\big)
\Big],
\end{split}
\end{equation}
where $\sigma(\cdot)$ denotes the sigmoid function and $\tilde{t}_{i,c}\!\in\!\{0,1\}$ is the augmented supervision target for class $c$.

By jointly optimizing these objectives, the model mitigates the effect of single-label exclusivity by encouraging semantically related categories to share representations, leading to more balanced and coherent learning under long-tailed distributions.

\section{Experiments}
\label{experiments} 

\subsection{Experimental Settings}
\noindent \textbf{Datasets.} 
We conduct experiments on four widely used long-tailed benchmarks: ImageNet-LT~\cite{liu2019large}, Places365-LT~\cite{liu2019large}, iNaturalist2018~\cite{van2018inaturalist}, and CIFAR100-LT~\cite{cao2019learning}.
ImageNet-LT contains 115.8K images from 1,000 classes, with per-class samples ranging from 5 to 1,280. Places365-LT includes 62.5K images across 365 categories, each containing 5–4,980 images. iNaturalist2018 consists of 437.5K images from 8,142 species, exhibiting a distribution from 2 to 1,000 images per class. CIFAR100-LT is a synthetic long-tailed version of CIFAR100 constructed with imbalance ratios (IR) of 100, 50, and 10.

% \vspace{0.5em}

\noindent \textbf{Metrics.}
Following the protocol of~\cite{liu2019large}, we evaluate all methods on a balanced test set and report the top-1 classification accuracy for three splits: Many-shot (more than 100 samples), Medium-shot (20 to 100 samples), and Few-shot (fewer than 20 samples), as well as the overall accuracy across all classes.

% \vspace{0.5em}

\noindent \textbf{Baselines.}
We compare our method with recent CLIP-based long-tailed recognition approaches, including LiVT~\cite{xu2023learning}, LPT~\cite{dong2023lpt}, VL-LTR~\cite{tian2022vl}, RAC~\cite{long2022retrieval}, BALLAD~\cite{ma2021simple}, and LIFT~\cite{shi2024long}. Following their standard setups, we ensure a fair comparison. Notably, VL-LTR and RAC leverage additional  data sources.

% \vspace{0.5em}

\noindent \textbf{Implementation Details.}
Unless otherwise specified, we adopt CLIP-ViT-B/16 as the backbone and employ AdaptFormer~\cite{chen2022adaptformer} for parameter-efficient fine-tuning. The number of CLIP-selected samples is fixed at $k=5$ for all comparisons.
% , while results under other $k$ settings are provided in the Appendix~\textcolor[HTML]{367DBD}{C}.
We use SGD with a learning rate of 0.01, momentum of 0.9, weight decay of $5\times10^{-4}$, batch size of 128, and a cosine learning rate schedule. All models are implemented in PyTorch and trained on a single NVIDIA GeForce RTX 5090 GPU.

\subsection{Comparison with State-of-the-art Methods}
\begin{table*}[!t]
\centering
\caption{Results on CIFAR100-LT comparing different \textit{Finetuning}-based methods under varying imbalance ratios. \textbf{Bold} indicates the best results. * marks reproduced results because the original paper did not report them completely.}
\label{tab:cifar100}
\footnotesize
\begin{tabular}{l|c|cccc|cccc|cccc}
\toprule
\multirow{2}{*}{\textbf{Method}} & \multirow{2}{*}{\textbf{Param.}} 
& \multicolumn{4}{c|}{\textbf{CIFAR100-IR100}}                                    
& \multicolumn{4}{c|}{\textbf{CIFAR100-IR50}}                                      
& \multicolumn{4}{c}{\textbf{CIFAR100-IR10}}                                    \\
& & \textbf{All} & \textbf{Many} & \textbf{Med.} & \textbf{Few} 
& \textbf{All} & \textbf{Many} & \textbf{Med.} & \textbf{Few} 
& \textbf{All} & \textbf{Many} & \textbf{Med.} & \textbf{Few} \\
\midrule
LiVT~\cite{xu2023learning} & 85.80M 
& 58.2 & - & - & - 
& - & - & - & - 
& 69.2 & - & - & - \\

BALLAD~\cite{ma2021simple} & 149.62M 
& 77.8 & 84.9 & 79.7 & 67.3
& - & - & - & -
& - & - & - & - \\

LIFT*~\cite{shi2024long} & 0.10M 
& 80.3 & 84.6 & \textbf{81.3} & 74.3
& 81.8 & 84.0 & 80.3 & 80.5
& 83.5 & 83.9 & 82.5 & - \\

\midrule
CUE (Ours) & 0.10M 
& \textbf{82.8} & \textbf{85.7} & 80.5 & \textbf{82.0}
& \textbf{83.6} & \textbf{86.0} & \textbf{81.0} & \textbf{83.9}
& \textbf{85.7} & \textbf{85.9} & \textbf{85.1} & - \\
\bottomrule
\end{tabular}
\vspace{-1em}
\end{table*}

\begin{table}[!t]
\centering
\caption{Results on ImageNet-LT.  \textbf{Bold} indicates the best results.}
\label{tab:imagenet}
\footnotesize
\begin{tabular}{l|c|ccccc}
\toprule
\textbf{Method}  & \textbf{Param.} & \textbf{All} & \textbf{Many} & \textbf{Med.} & \textbf{Few} \\
\midrule
LiVT~\cite{xu2023learning}    & 85.80M              & 60.9          & 73.6          & 56.4          & 41.0          \\
Decoder~\cite{wang2024exploring} & 21.26M         & 73.2          & -             & -             & -             \\
BALLAD~\cite{ma2021simple}  & 149.62M        & 75.7          & 79.1          & 74.5          & 69.8          \\
LIFT~\cite{shi2024long}    & \textbf{0.62M} & 77.0          & 80.2          & 76.1          & 71.5          \\
\midrule
CUE (Ours)    & \textbf{0.62M} & \textbf{77.4} & \textbf{80.3} & \textbf{76.3} & \textbf{73.0} \\
\bottomrule
\end{tabular}
\end{table}

\begin{table}[!t]
\centering
\caption{Results on Places-LT. \textbf{Bold} indicates the best results.}
\label{tab:places}
\footnotesize
\begin{tabular}{l|c|ccccc}
\toprule
\textbf{Method}  & \textbf{Param.} & \textbf{All} & \textbf{Many} & \textbf{Med.} & \textbf{Few} \\
\midrule
LiVT~\cite{xu2023learning}    & 85.80M               & 40.8          & 48.1          & 40.6          & 27.5          \\
Decoder~\cite{wang2024exploring} & 21.26M           & 46.8          & -             & -             & -             \\
RAC~\cite{long2022retrieval}     & 85.80M           & 47.2          & 48.7          & 48.3          & 41.8          \\
BALLAD~\cite{ma2021simple}       & 149.62M          & 49.5          & 49.3          & 50.2          & 48.4          \\
VL-LTR~\cite{tian2022vl}                          & 149.62M          & 50.1          & \textbf{54.2} & 48.5          & 42.0          \\
LPT~\cite{dong2023lpt}           & 1.01M            & 50.1          & 49.3          & 52.3          & 46.9          \\
LIFT~\cite{shi2024long}          & \textbf{0.18M}   & 51.5          & 51.3          & \textbf{52.2}          & 50.5          \\
\midrule
CUE (Ours)                       & \textbf{0.18M}   & \textbf{51.7} & 50.6          & \textbf{52.2} & \textbf{52.4} \\
\bottomrule
\end{tabular}
\end{table}

We evaluate CUE on four benchmarks of distinct characteristics, as summarized in Tabs.~\ref{tab:cifar100}-\ref{tab:inat}, covering synthetic, standardized, and naturally long-tailed scenarios.

% \vspace{0.5em}

\noindent \textbf{Synthetic small-scale benchmarks.}
On the small-scale CIFAR100-LT, which allows customized imbalance settings, CUE consistently outperforms existing methods across all imbalance ratios. Under the most challenging IR100 condition, it achieves up to \textbf{+7.7\%} improvement on the Few-shot categories and overall gains of \textbf{+2.5\%}, \textbf{+1.8\%}, and \textbf{+2.2\%} for IR100, IR50, and IR10, respectively. These improvements primarily stem from stronger generalization on under-represented classes while maintaining comparable performance on head and medium categories. It is worth noting that the results of LIFT are not reported in its original paper; thus, we reproduced them under the same experimental environment for fairness.

% \vspace{0.5em}

\noindent \textbf{Standardized large-scale benchmarks.}
On ImageNet-LT and Places-LT, where the training and evaluation splits follow established long-tailed protocols~\cite{liu2019large}, CUE achieves the most balanced performance among finetuning-based approaches. On ImageNet-LT, it attains a \textbf{+1.5\%} improvement on Few-shot classes and an overall accuracy of \textbf{77.4\%} (\textbf{+0.4\%} over LIFT), highlighting its ability to preserve pre-trained semantics during adaptation. Similarly, on Places-LT, CUE improves Few-shot accuracy by \textbf{+1.9\%} and achieves \textbf{51.7\%} overall accuracy (\textbf{+0.2\%} over LIFT), demonstrating balanced performance in large-scale scene recognition.

% \vspace{0.5em}

\noindent \textbf{Naturally long-tailed benchmarks.}
On iNaturalist 2018, which exhibits an inherent real-world long-tailed distribution, CUE achieves a more globally balanced representation across categories.
While LIFT suffers from head-class degradation due to over-adaptation, CUE reverses this trend through concept-aware multi-label expansion, improving the Many and Few categories by \textbf{+1.0\%} and \textbf{+0.6\%}, respectively, and reaching an overall accuracy of \textbf{79.6\%} (\textbf{+0.5\%} over LIFT).
These results demonstrate that CUE generalizes well to naturally imbalanced real-world data.

\begin{table}[!t]
\centering
\caption{Results on iNaturalist. \textbf{Bold} indicates the best results.}
\label{tab:inat}
\footnotesize
\begin{tabular}{l|c|ccccc}
\toprule
\textbf{Method}  & \textbf{Param.} & \textbf{All} & \textbf{Many} & \textbf{Med.} & \textbf{Few} \\
\midrule
LiVT~\cite{xu2023learning}    & 85.80M              & 76.1          & 78.9          & 76.5          & 74.8          \\
Decoder~\cite{wang2024exploring} & 21.26M         & 59.2          & -             & -             & -             \\
VL-LTR~\cite{tian2022vl}  & 149.62M        & 76.8          & -             & -             & -             \\
LPT~\cite{dong2023lpt}   & \textbf{1.01M} & 76.1          & -             & -             & 79.3          \\
LIFT~\cite{shi2024long}  & 4.75M          & 79.1          & 72.4          & 79.0          & 81.1          \\
\midrule
CUE (Ours)               & 4.75M          & \textbf{79.6} & \textbf{73.4} & \textbf{79.2} & \textbf{81.7} \\
\bottomrule
\end{tabular}
\vspace{-2em}
\end{table}

\subsection{Ablation Studies}
\noindent \textbf{Effect of Model Components.}
We perform an ablation study to verify the effectiveness of each component in CUE under long-tailed fine-tuning.
Starting from the \textbf{LIFT} baseline (AdaptFormer with Logit Adjustment Loss), we progressively incorporate the VLM-based instance-level cues and the LLM-based class-level cues.
As summarized in Table~\ref{tab:ablation}, both cues yield consistent and noticeable gains across multiple datasets, with the most significant improvements observed in the tail (\textit{Few-shot}) categories.
The VLM-based cue generally provides slightly higher gains than the LLM-based one, suggesting that instance-level visual guidance offers more direct and fine-grained supervision during adaptation.
When combined, the full CUE model achieves the most balanced and stable performance, indicating that the two cues provide complementary benefits by jointly enhancing feature coherence and semantic consistency.
% To evaluate the effectiveness of each component in our proposed CUE framework, we conduct an ablation study on CIFAR100-IR100, ImageNet-LT, and Places-LT. Starting from the baseline model trained with the standard Logit-Adjusted (LA) loss, we progressively introduce the VLM-based instance-level prior and the LLM-based class-level prior. As shown in Table~\ref{tab:ablation}, both components independently bring clear improvements, particularly benefiting the tail (\textit{Few}) categories. When combined, the full CUE model achieves the best overall balance across all data regimes. Although the two cues partially address related types of semantic confusion, their integration still provides consistent gains, indicating that they offer complementary yet reinforcing guidance for preserving inter-class coherence and mitigating fine-grained confusion under long-tailed distributions.

\begin{table*}[!t]
\centering

\caption{Ablation of CUE components on three datasets. Both cues bring consistent improvements, particularly on the tail categories. \textbf{Bold} indicates the best results.
}
\label{tab:ablation}
\small
\begin{tabular}{cc|cccc|cccc|cccc}
\toprule
\multirow{2}{*}{VLM} & \multirow{2}{*}{LLM} &
\multicolumn{4}{c|}{CIFAR100-IR100} & \multicolumn{4}{c|}{ImageNet-LT} & \multicolumn{4}{c}{Places-LT} \\
\cmidrule(lr){3-6} \cmidrule(lr){7-10} \cmidrule(lr){11-14}
 &  & \textbf{All} & \textbf{Many} & \textbf{Med.} & \textbf{Few} & \textbf{All} & \textbf{Many} & \textbf{Med.} & \textbf{Few} & \textbf{All} & \textbf{Many} & \textbf{Med.} & \textbf{Few} \\
\midrule
 &  & 80.3  & 84.6 & 81.3  & 74.3 & 77.0 & 80.2 & 76.1 & 71.5   & 51.5 & \textbf{51.3} & 52.2 & 50.5 \\
\checkmark &  & \textbf{82.8} & 85.7 & 80.8 & 81.6  & \textbf{77.4} & \textbf{80.4} & \textbf{76.3} & 72.6 & 51.6 & 50.5 & \textbf{52.3} & 52.2 \\
 & \checkmark & 82.7 & \textbf{86.3} & \textbf{81.3} & 80.2 & 77.3 & \textbf{80.4} & \textbf{76.3} & 72.5 & 51.5 & 50.7 & \textbf{52.3} & 51.1 \\
\checkmark & \checkmark & \textbf{82.8} & 85.7 &	80.5 & \textbf{82.0}	 & \textbf{77.4} & 80.3 & \textbf{76.3} & \textbf{73.0} & \textbf{51.7} & 50.6 & 52.2 & \textbf{52.4} \\
\bottomrule
\end{tabular}
\vspace{-1.0em}
\end{table*}

% \vspace{0.5em}

\begin{figure}[!t]
\centering
\setlength{\tabcolsep}{3pt}
\begin{tabular}{cc}
\includegraphics[width=0.49\linewidth]{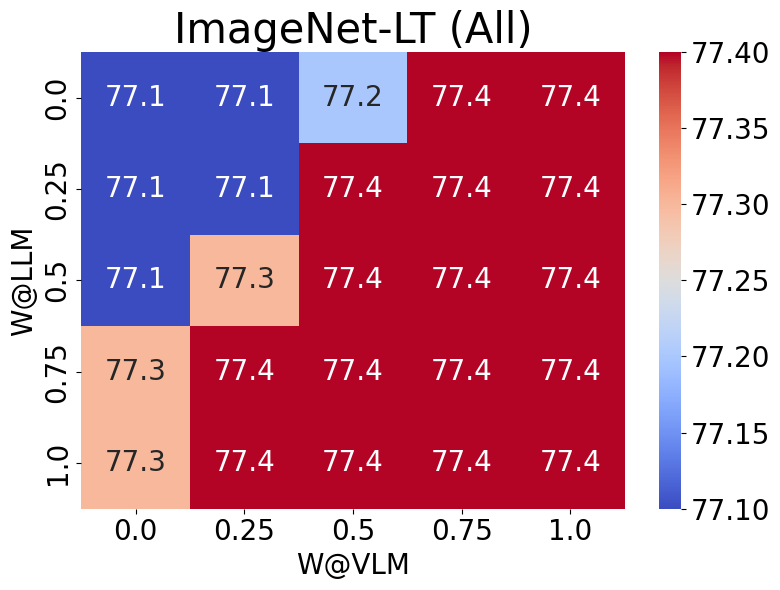} &
\includegraphics[width=0.49\linewidth]{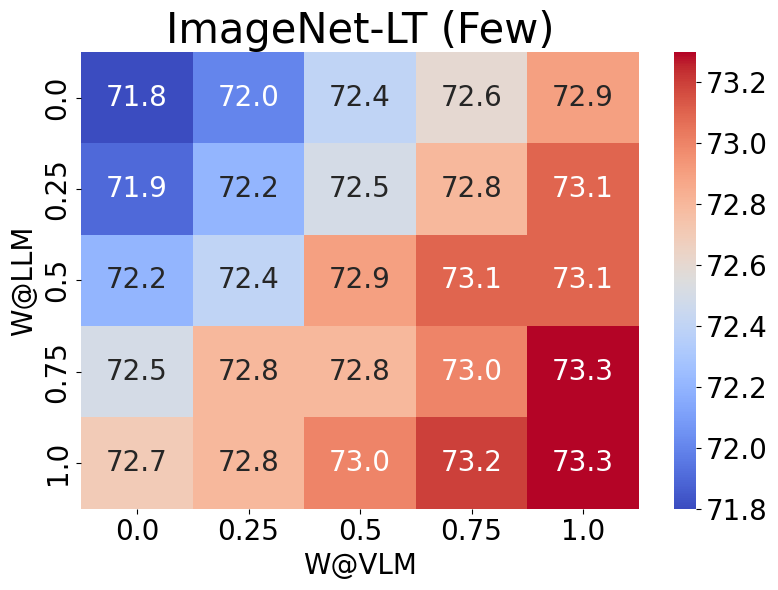} \\
\includegraphics[width=0.49\linewidth]{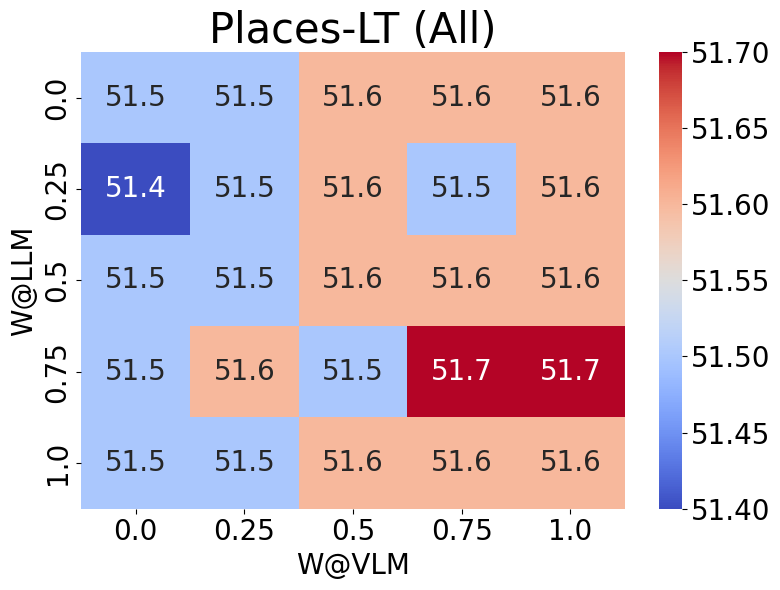} &
\includegraphics[width=0.49\linewidth]{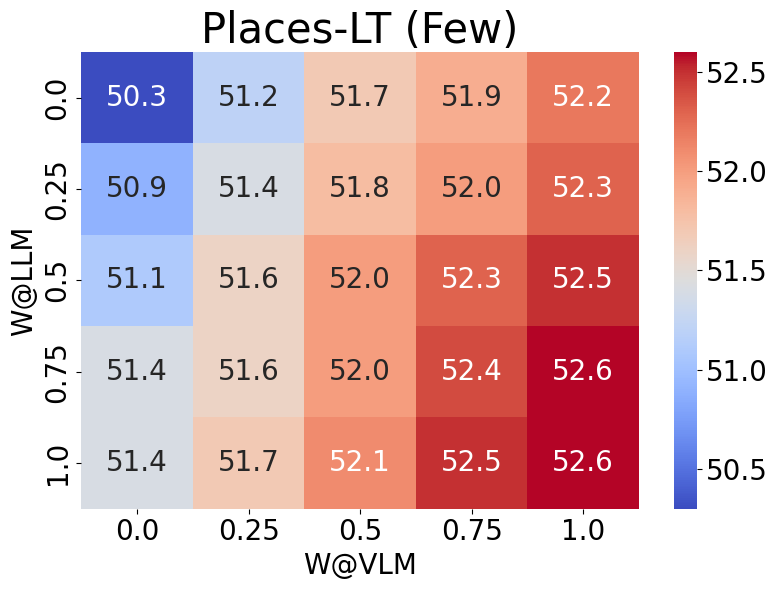}
\end{tabular}
\caption{Sensitivity analysis of CUE with respect to the weighting coefficients $(\lambda_{\text{VLM}}, \lambda_{\text{LLM}})$ on ImageNet-LT and Places-LT (avg/few). 
}
\vspace{-2em}
\label{fig:hyper_sensitivity}
\end{figure}

\noindent \textbf{Hyperparameter Sensitivity.}
% We further investigate the sensitivity of CUE to the weighting coefficients $\lambda_{\text{VLM}}$ and $ \lambda_{\text{LLM}}$. For ImageNet-LT and Places-LT, each coefficient is varied from 0.0 to 1.0 with a step size of 0.25. 
We further examine the sensitivity of CUE to the weighting coefficients $\lambda_{\text{VLM}}$ and $\lambda_{\text{LLM}}$ on ImageNet-LT and Places-LT, varying each from 0.0 to 1.0 with a step size of 0.25. As shown in Fig.~\ref{fig:hyper_sensitivity}, CUE demonstrates strong robustness across a wide range of parameter settings. Specifically, incorporating either VLM or LLM guidance does not lead to performance degradation; with appropriate weighting, both contribute to consistent improvements. Configurations using VLM-only generally outperform those using LLM-only, while combining both cues (non-zero $\lambda_{\text{VLM}}$ and $\lambda_{\text{LLM}}$) often yields the best results. Across all settings, clear gains are observed on tail classes. However, excessive reliance on either cue can slightly reduce accuracy, implying that visual and semantic cues, while synergistic, play distinct roles and should be balanced to effectively mitigate concept confusion in long-tailed learning.

\subsection{Extension Experiments}
To examine the generality and applicability of CUE, we evaluate it under different adaptation paradigms, including parameter-efficient fine-tuning and from-scratch long-tailed training. 
For fair comparison, we follow the standard settings of prior work for each paradigm.
All experiments in this section are conducted on CIFAR100-LT with an imbalance ratio of IR=100. 
% for consistency and computational efficiency.
% Results on additional datasets are provided in Appendix~\textcolor[HTML]{367DBD}{D}.

\begin{figure*}[!t]
\centering
\includegraphics[width=1.0\linewidth]{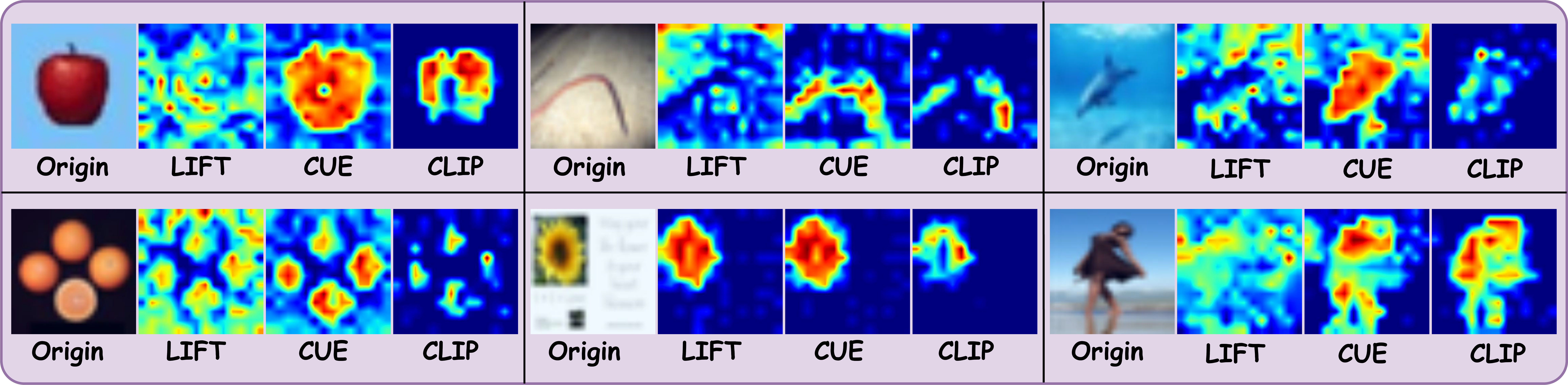}
\vspace{-1.7em}
\caption{\textbf{Grad-CAM visualization comparing LIFT, CUE and CLIP on several examples}. By sharing related  cues across correlated categories, CUE produces more precise and coherent localization than both LIFT and CLIP.}
\vspace{-1.7em}
\label{fig:gradcam}
\end{figure*}

\label{sec:extension}
% \vspace{0.5em}

\noindent \textbf{Extension to PEFT Methods.}
To examine the generality and flexibility of CUE, we evaluate its integration with several representative parameter-efficient fine-tuning (PEFT) strategies, including LoRA~\cite{hu2022lora}, VPT-shallow~\cite{jia2022visual}, VPT-deep~\cite{jia2022visual}, Adapter~\cite{houlsby2019parameter}, and AdaptFormer~\cite{chen2022adaptformer}.
As reported in Table~\ref{tab:extension_peft}, CUE consistently enhances the performance of all PEFT backbones on CIFAR100-LT (IR=100).
Notably, the gains are particularly pronounced for tail categories, indicating that CUE generalizes well across different parameter-efficient fine-tuning strategies.

\begin{table}[!t]
\centering
\caption{Extension of CUE to different parameter-efficient fine-tuning (PEFT) methods on CIFAR100-IR100. CUE consistently improves performance across various               strategies.}
\label{tab:extension_peft}
\small
\begin{tabular}{l|cccc}
\toprule
\textbf{Method} & \textbf{All} & \textbf{Many} & \textbf{Med.} & \textbf{Few} \\
\midrule
LoRA~\cite{hu2022lora} & 79.1 & 84.0 & \textbf{79.7} & 72.7 \\
+ CUE & \textbf{81.2} & \textbf{84.8} & 79.0 & \textbf{79.8} \\
\midrule
VPT-shallow~\cite{jia2022visual} & 69.7 & \textbf{79.4} & 74.9 & 52.2 \\
+ CUE & \textbf{76.5} & 78.2 & \textbf{75.9} & \textbf{75.2} \\
\midrule
VPT-deep~\cite{jia2022visual} & 78.7 & 81.9 & \textbf{80.0} & 73.5 \\
+ CUE & \textbf{81.0} & \textbf{83.3} & 78.9 & \textbf{80.8} \\
\midrule
Adapter~\cite{houlsby2019parameter} & 79.6 & 83.7 & 80.9 & 73.5 \\
+ CUE & \textbf{82.9} & \textbf{85.8} & 80.9 & \textbf{82.0} \\
\midrule
AdaptFormer~\cite{chen2022adaptformer} & 80.3 & 84.6 & \textbf{81.3} & 74.3 \\
+ CUE & \textbf{82.8} & \textbf{85.7} & 80.5 & \textbf{82.0} \\

\bottomrule
\end{tabular}
\vspace{-1.5em}
\end{table}

% 2. 扩展到from scratch
\vspace{0.5em}

\noindent \textbf{Extension to From-Scratch Training.}
We further verify that CUE is not limited to fine-tuning of pre-trained models but can also be applied to conventional long-tailed training from scratch. 
Since from-scratch training also adopts single-label supervision, the inherent mutual exclusivity still leads to semantic suppression among related categories, resulting in concept confusion. 
Given the plug-and-play nature of CUE, we directly integrate it into several representative from-scratch methods, including DODA~\cite{wang2024kill}, LOS~\cite{sun2025rethinking}, LA~\cite{menon2021longtail}, and ResLT~\cite{cui2022reslt}, without modifying their architectures or optimization pipelines. 
As summarized in Table~\ref{tab:extension_scratch}, all baselines benefit consistently from the inclusion of CUE, demonstrating its generality in mitigating concept confusion across different long-tailed learning paradigms.

\begin{table}[!t]
\centering
\caption{Extension of CUE to from-scratch long-tailed learning methods on CIFAR100-IR100. Consistent improvements demonstrate the plug-and-play nature of our approach.}
\vspace{-0.5em}
\label{tab:extension_scratch}
\small
\begin{tabular}{l|cccc}
\toprule
Method & \textbf{All} & \textbf{Many} & \textbf{Med.} & \textbf{Few} \\
\midrule
DODA~\cite{wang2024kill} &  45.7  &  \textbf{62.1}  &  44.8  &  28.1  \\
+ CUE &  \textbf{47.3}  &  61.8  &  \textbf{45.3}  &  \textbf{33.2}  \\
\midrule
LOS~\cite{sun2025rethinking} &  26.6  &  \textbf{60.7}  &  14.3  &  1.5  \\
+ CUE &  \textbf{31.9}  &  60.2  &  \textbf{25.6}  &  \textbf{6.8}  \\
\midrule
LA~\cite{menon2021longtail} & 42.9 & \textbf{61.3} & 41.0 & 24.1 \\
+ CUE & \textbf{44.1} & 58.8 & \textbf{41.8} & \textbf{29.9} \\
\midrule
ResLT~\cite{cui2022reslt} & 40.4 & \textbf{58.2} & 42.1 & 18.6 \\
+ CUE & \textbf{43.7} & 55.7 & \textbf{44.1} & \textbf{29.6} \\
\bottomrule
\end{tabular}
\vspace{-1.5em}
\end{table}

\subsection{Additional Strengths and Analyses}
\noindent \textbf{(1) Balanced Feature Learning.}
To further understand how CUE improves class-wise balance, we examine the mean number of misclassified samples for \textit{Many}, \textit{Medium}, and \textit{Few} categories, along with the overall balancedness of feature representations. 
Following the balance metric $\beta(V)$ with $\sigma=0.1$ from \cite{kang2020exploring}, we measure feature uniformity based on the pairwise similarity of class-wise accuracies, where a higher $\beta(V)$ indicates more uniform separability across classes. 
As shown in Table~\ref{tab:balance}, CUE consistently achieves fewer misclassifications and higher balancedness scores than LIFT across all three datasets, demonstrating that CUE promotes more balanced and unbiased feature learning by mitigating concept confusion. 
This observation echoes the pattern in Fig.~\ref{fig:motivation2}(a), where baseline models exhibit frequent tail-class misclassifications caused by concept confusion, while CUE effectively alleviates this issue and yields more balanced results.

\begin{table}[!t]
\centering
\caption{Comparison of class-wise balance between LIFT and CUE on three datasets. 
We report mean number of misclassified after fine-tuning for head (\textit{Many}), medium  (\textit{Med.}), and tail (\textit{Few}) classes, 
along with the balancedness score (higer $\beta(V)$ indicates more uniform separability).}
\vspace{-0.5em}
\label{tab:balance}
\footnotesize
\begin{tabular}{l|cccc}
\toprule
\textbf{Method} & \textbf{Many} & \textbf{Med.} & \textbf{Few} & $\beta(V)$\textbf{ ($\uparrow$)} \\
\midrule
\textcolor{gray}{\textbf{CIFAR100-IR100}} & & & & \\
\quad LIFT & 3.1 & \textbf{3.2} & 8.6 & 77.5\% \\
\quad CUE (ours) & \textbf{2.6} & 3.6 & \textbf{4.6} & \textbf{82.9\%} \\[3pt]
\midrule
\textcolor{gray}{\textbf{ImageNet-LT}} & & & & \\
\quad LIFT & 1.9 & 2.6 & 3.7 & 69.7\% \\
\quad CUE (ours) & 1.9 & 2.6 & \textbf{3.2} & \textbf{70.9\%} \\[3pt]
\midrule
\textcolor{gray}{\textbf{Places-LT}} & & & & \\
\quad LIFT & \textbf{5.7} & 5.4 & 7.9 & 60.4\% \\
\quad CUE (ours) & 5.8 & \textbf{5.3} & \textbf{6.9} & \textbf{60.8\%} \\

\bottomrule
\end{tabular}
\vspace{-1.5em}
\end{table}

% gradcam可视化展示 [突出我们的优势]

% \vspace{0.5em}

\noindent \textbf{(2) Improved Feature Localization.}
To further analyze the attention behavior, we visualize Grad-CAM maps for LIFT, CUE, and zero-shot CLIP. As shown in Fig.~\ref{fig:gradcam}, LIFT often focuses on incorrect or irrelevant regions due to concept confusion, while CLIP localizes objects more accurately but lacks domain adaptation to the target dataset. In contrast, CUE inherits the object-centric property of CLIP and enhances it through multi-label cues that encourage related categories to share semantic guidance, resulting in more precise and coherent feature localization.

% \vspace{0.5em}

\noindent \textbf{(3) Effective  Multi-Label Relations.}
% 打标签的相关性示例以及反向实验
% Finally, we present examples of multi-labels derived from the VLM- and LLM-based cues to demonstrate their semantic coherence. The detailed label mappings are provided in Appendix~\textcolor[HTML]{367DBD}{E}. 
To examine the reliability of these cues, we replace the top-$k$ VLM-selected samples with random or last-$k$ ones. As shown in Table~\ref{tab:multi_label_relation}, this substitution leads to a clear degradation in performance, confirming that the proposed cues capture meaningful and semantically consistent relationships rather than spurious correlations.

\begin{table}[!t]
\centering
\caption{Effect of multi-label relation cues on CIFAR100-LT (IR=100). Replacing top-$k$ samples in the VLM-based cue selection with random-$k$ or last-$k$ ones leads to a clear performance drop, verifying the semantic effectiveness of the proposed cues.}
\vspace{-0.5em}
\label{tab:multi_label_relation}
\begin{tabular}{l|cccc}
\toprule
\textbf{Method} & \textbf{All} & \textbf{Many} & \textbf{Med.} & \textbf{Few} \\
\midrule
VLM-based (Top-$k$) & \textbf{82.8} & \textbf{85.7} & 80.8 & \textbf{81.6} \\
VLM-based (random-$k$) & 80.1 & 83.6 & \textbf{81.1} & 75.0 \\
VLM-based (Last-$k$) & 79.5 & 84.0 & 80.9 & 72.5 \\
\bottomrule
\end{tabular}
\vspace{-1.5em}
\end{table}

\section{Conclusion}
\label{conclusion}
We introduced CUE, a framework that alleviates concept confusion in long-tailed learning by leveraging multi-label cues derived from pre-trained vision–language and language models. 
By encouraging related categories to share semantic guidance, CUE preserves inter-class relationships during adaptation and achieves more balanced recognition across head and tail classes. 
Its plug-and-play design allows flexible integration into both fine-tuning and from-scratch training paradigms. While effective, its performance still depends on the quality of external cues, suggesting future work on adaptive cue generation and lightweight integration for large-scale applications.

\sloppy
\section*{Acknowledgment}
This study was supported in part by the National Natural Science Foundation of China under Grants 62376233, 62506
270, 62502402 and 62306181; in part by the Natural Science Foundation of Fujian Province under Grant 2024J090
01; in part by the Guangdong Basic and Applied Basic Research Foundation under Grant 2024A1515010163; in part by the Shenzhen Science and Technology Program under Grant RCBS20231211090659101; in part by the National Key Laboratory of Radar Signal Processing under Grant JKW202403; and in part by Xiaomi Young Talents Program.
\fussy
{
    \small
    \bibliographystyle{ieeenat_fullname}
    \bibliography{main}
}

% WARNING: do not forget to delete the supplementary pages from your submission 
% \input{sec/X_suppl}

\end{document}